\documentclass[times, review, 10pt]{elsarticle}

\usepackage{amssymb}
\usepackage{amsmath}
\usepackage{booktabs}
\usepackage{makecell,array}
\usepackage{graphicx}
\usepackage{subcaption}
\usepackage{hyperref}
\usepackage{siunitx}
\usepackage{multirow}
\usepackage{adjustbox}
\usepackage{enumitem}
\usepackage{xcolor}

\newcolumntype{Y}{>{\raggedright\arraybackslash}X}

\journal{Pattern Recognition}

\begin{document}
\begin{frontmatter}
\title{Interaction as Interference: A Quantum-Inspired Aggregation Approach\tnoteref{t1}}
\tnotetext[t1]{Preprint submitted to \textit{Pattern Recognition}, November 2025.}
\author{Pilsung Kang}
\ead{pilsungk@dankook.ac.kr}
\affiliation{organization={Department of Software Science, Dankook University},
city={Yongin}, postcode={16890}, country={South Korea}}

%% \title{Interaction as Interference: A Quantum-Inspired Aggregation Approach}
%% %% \tnotetext[label1]{}
%% \author{Pilsung Kang}
%% \ead{pilsungk@dankook.ac.kr}
%% \affiliation{organization={Department of Software Science, Dankook University}, city={Yongin}, postcode={16890}, country={South Korea}}
%% 
\begin{abstract}
Classical approaches often treat interaction as engineered product terms or as emergent patterns in flexible models, offering little control over how synergy or antagonism arises. We take a quantum-inspired view: following the Born rule (probability as squared amplitude), \emph{coherent} aggregation sums complex amplitudes before squaring, creating an interference cross-term, whereas an \emph{incoherent} proxy sums squared magnitudes and removes it. In a minimal linear-amplitude model, this cross-term equals the standard potential-outcome interaction contrast \(\Delta_{\mathrm{INT}}\) in a \(2\times 2\) factorial design, giving relative phase a direct, mechanism-level control over synergy versus antagonism.

We instantiate this idea in a lightweight \emph{Interference Kernel Classifier} (IKC) and introduce two diagnostics: \emph{Coherent Gain} (log-likelihood gain of coherent over the incoherent proxy) and \emph{Interference Information} (the induced Kullback–Leibler gap). A controlled phase sweep recovers the identity. On a high-interaction synthetic task (XOR), IKC outperforms strong baselines under paired, budget-matched comparisons; on real tabular data (\emph{Adult} and \emph{Bank Marketing}) it is competitive overall but typically trails the most capacity-rich baseline in paired differences. Holding learned parameters fixed, toggling aggregation from incoherent to coherent consistently improves negative log-likelihood, Brier score, and expected calibration error, with positive Coherent Gain on both datasets. 
\end{abstract}

%% Keywords

\end{frontmatter}


\section{Introduction}\label{s:intro}

Interaction effects are of central importance across science and machine learning. They arise when the combined influence of multiple factors departs from simple additivity—producing synergy, where effects amplify each other, or antagonism, where they suppress one another. For example, combining two drugs can lead to markedly beneficial or adverse outcomes depending on their interaction. Yet frameworks for understanding interaction remain incomplete. In causal inference, interactions are commonly introduced as ad-hoc product terms in a regression, describing the phenomenon without explaining its generative mechanism~\cite{pearl:2009:causality,spirtes:2000:book}. Likewise, classical information measures such as mutual information quantify statistical dependence but do not reveal how multiple inputs compose—e.g., they do not distinguish synergy from redundancy nor encode the sign of interaction. In contrast, a foundational principle of quantum theory, the Born rule, posits that observable probabilities are squared amplitudes~\cite{nielsen2010quantum}; this coherent (phase-sensitive) aggregation naturally produces interference cross-terms that can amplify or attenuate outcome probabilities. The gap between ``interaction as a statistical add-on'' and ``interaction as a property of an aggregation rule'' motivates our quantum-inspired perspective on causal interaction.

In this work, we propose that the origin of interaction lies not in an extra product term but in the rule by which causal influences are aggregated. We recast interaction as a consequence of coherent aggregation in the sense of the Born rule. To build intuition, think of two ripples on water: alignment produces amplification (synergy) and misalignment produces cancellation (antagonism). In our framework, each influence is represented by a complex amplitude \(\psi\) (a pre-probability score); observable probability is the squared magnitude \(|\psi|^{2}\) per the Born rule. With two binary causes \(A,B\in\{0,1\}\), we write the cell probability \(p_{ab}\equiv P(Y{=}1\mid A{=}a,B{=}b)\) and define the interaction contrast
\[
\Delta_{\mathrm{INT}}\;\equiv\;p_{11}-p_{10}-p_{01}+p_{00}.
\]
When these causes are randomized or well controlled, this contrast equals the potential-outcome interaction contrast; we use the \(p\)-based definition 
throughout for clarity.

To formalize aggregation, we distinguish two rules. The incoherent proxy adds energies (squared magnitudes), \(p_{\mathrm{inc}}=\sum_k |\psi_k|^2\). By contrast, coherent aggregation adds amplitudes before squaring, \(p_{\mathrm{coh}}=\bigl|\sum_k \psi_k\bigr|^2\), which introduces an interference cross-term. In a minimal amplitude-linear model for the \(2\times2\) design,
\[
\psi(a,b)=u_0+a\,u_A+b\,u_B,\qquad u_0,u_A,u_B\in\mathbb{C},\quad p_{ab}=|\psi(a,b)|^2,
\]
the interaction contrast reduces exactly to the cross-term
\[
\Delta_{\mathrm{INT}} \;=\; 2\,\mathrm{Re}\!\left(u_A\,u_B^{*}\right),
\]
making the relative phase between \(u_A\) and \(u_B\) a direct geometric control of synergy versus antagonism. (A short proof appears in Methods~\ref{s:method}.) This identity establishes $\Delta_{\mathrm{INT}}$ as a direct observable signature of the aggregation rule itself.

Our contributions are thus conceptual, methodological, and empirical. We formalize interaction as a property of a phase-sensitive aggregation rule, instantiate it in a practical classifier (the Interference Kernel Classifier, or IKC), and introduce two diagnostic metrics to quantify its effect: \emph{Coherent Gain} ($G_{\mathrm{coh}}$), the per-instance log-likelihood gain over an interaction-free incoherent proxy, and \emph{Interference Information} ($\mathcal{J}_{\mathrm{int}}$), the Kullback-Leibler (KL) divergence between the coherent and incoherent conditional distributions.  Through a series of experiments—from controlled phase sweeps that validate the core identity to applications on high-interaction tasks (XOR) and real-world data (\emph{Adult} and \emph{Bank Marketing}) that demonstrate improved calibration and robustness—we validate our framework. Our approach is classical in implementation and makes no quantum-advantage claim; it provides a practical toolkit and a unifying lens for dissecting and engineering interaction in causal learning.

The remainder of this paper is organized as follows. Section~\ref{s:related} situates our work relative to prior research on interaction modeling. Section~\ref{s:method} develops our theoretical framework, formalizes the interference-interaction identity, and details the proposed model. Section~\ref{s:experiments} describes the protocols for our five empirical studies, and Section~\ref{s:results} presents their findings. Finally, we discuss the implications and limitations of our work in Section~\ref{s:discuss} and conclude in Section~\ref{s:conc}.

%%%%%%%%%%%%%%%%%%%%%%%%%%%5%%%%%%%%%%%%%%%%%%%%%%%%%%%%%%%%%%%%%%%%%%%
\section{Related Work}\label{s:related}

This section situates our contribution within three strands: (i) classical interaction modeling in statistics and machine learning (ML), (ii) causal definitions of interaction in the potential-outcome and structural frameworks, and (iii) quantum-inspired aggregation alongside specialized ML architectures that capture interactions. Prior approaches typically either enumerate interaction terms or recover them implicitly, offering limited mechanism-level control over synergy versus antagonism. In contrast, we frame interaction as an aggregation principle: a learnable phase governs synergy versus antagonism; the concrete mechanism and its connection to the potential-outcome interaction contrast \(\Delta_{\mathrm{INT}}\) are developed later in the paper.

\subsection{Classical Interaction Modeling}

Generalized Additive Models (GAMs) estimate nonlinear main effects additively, and in their basic form exclude interactions~\cite{hastie1986,wood:2017:gam}. Interactions can be added via tensor-product smooths or explicit products, but this requires pre-selecting orders and strong regularization, so apparent interaction strength can be an artifact of modeling choices. Post-hoc diagnostics such as the H\,-statistic quantify interaction strength in tree ensembles~\cite{friedman:2008:hstat}, yet they assess rather than control interactions. By contrast, our approach treats interaction as a property of the aggregation rule: coherent (phase-sensitive) addition yields an interference cross-term whose sign and magnitude are governed by a learned phase, providing mechanism-level control.

Tree ensembles, especially gradient boosting~\cite{friedman:2001:gbm}, capture high-order interactions via recursive partitioning and additive expansions. Yet these interactions are emergent, not tunable, components—offering limited mechanism-level control or interpretation compared to our phase parameter that directly sets the sign and strength of the interference term.

Hierarchical interaction selection imposes heredity constraints so interactions appear only with main effects (e.g., hierarchical lasso~\cite{bien:2013}); it still requires prespecifying interaction order and primarily induces sparsity rather than a mechanism that distinguishes synergy from antagonism. Our approach instead encodes interaction in the aggregation rule itself, tying the interference cross-term directly to the potential-outcome contrast \(\Delta_{\mathrm{INT}}\) without enumerating pairs.

Rule ensembles make interactions explicit as sparse conjunctions extracted from trees, improving interpretability over black-box boosting, yet these terms are discovered indirectly from recursive splits and remain difficult to tune mechanistically~\cite{friedman:2008:hstat}. In our framework, the coherent--incoherent toggle manipulates only the aggregation step (holding parameters fixed), providing mechanism-level control and diagnostics \(\{G_{\mathrm{coh}},\mathcal{J}_{\mathrm{int}}\}\) rather than post-hoc identification of interaction structure.

\subsection{Causal Interaction in Potential-Outcome and SCM Frameworks}

In the potential-outcome framework with two binary causes \(A,B\in\{0,1\}\) and potential outcomes \(Y(a,b)\), the additive-scale interaction contrast \(\Delta_{\mathrm{INT}}\) quantifies synergy and antagonism as a contrast on cell probabilities (formal definition in Sec.~\ref{ssec:identity}).  Under randomized or well-controlled assignment (no unmeasured confounding), this contrast is identifiable from observed conditional probabilities.  Prior works~\cite{rubin:1974,vanderweele_knol:2014,vanderweele:2015} clarify interpretability on additive vs.\ multiplicative scales and provide conditions under which \(\Delta_{\mathrm{INT}}\) is identifiable.

From the structural causal model (SCM) viewpoint, interaction corresponds to non-additive composition in the structural equation for \(Y\)~\cite{pearl:2009:causality} and relates to sufficient-cause (synergy or antagonism) notions and to effect decompositions on additive or multiplicative scales. Our use of \(\Delta_{\mathrm{INT}}\) follows these definitions and aligns notation with the rest of the paper. Building on these definitions, we later show (Sec.~\ref{ssec:identity}) that in a minimal coherent aggregation model the interference cross-term equals \(\Delta_{\mathrm{INT}}\) exactly in a \(2{\times}2\) design, making the relative phase a tunable mechanism for the sign and magnitude of interaction (instantiated in Sec.~\ref{ssec:ikc}).

\subsection{Quantum-Inspired Aggregation and Specialized Interaction Models}

Quantum cognition has long used interference---adding amplitudes before squaring---to explain context and order effects in human judgment~\cite{busemeyer2012,pothos2022}. These models typically target explanatory adequacy on behavioral tasks and do not connect to causal interaction formalizations. By contrast, we cast interaction as an aggregation principle: coherent (phase-sensitive) addition yields an interference cross-term whose sign and magnitude are governed by a learned phase, and we link this term exactly to the potential-outcome contrast \(\Delta_{\mathrm{INT}}\), delivering a practical classifier (IKC) with paired diagnostics \(\{G_{\mathrm{coh}},\mathcal{J}_{\mathrm{int}}\}\).

Specialized ML models also capture interactions but with different goals and mechanisms. Factorization Machines efficiently parameterize pairwise feature interactions in sparse settings~\cite{rendle2010factorization}; Wide\&Deep combines memorization and generalization for recommendation~\cite{cheng:2016:wide_deep}; Neural FM and xDeepFM learn higher-order interactions via neural parameterizations~\cite{he:2017:nfm,lian:2018:xdeepfm}; attention mechanisms learn data-dependent interaction weights across tokens~\cite{vaswani:2017:nips}. These approaches excel at prediction yet offer limited mechanism-level control over whether interactions are synergistic or antagonistic: the effect emerges from fitted parameters rather than a tunable aggregation rule. In IKC, a phase parameter directly controls the interference sign, and a coherent–incoherent toggle (holding parameters fixed) isolates the contribution of phase-coherent aggregation from representation capacity. For evaluation, we also employ standard post-hoc calibration via temperature scaling (TS)~\cite{guo:2017:calibration}, which complements but is orthogonal to the aggregation mechanism.

%%%%%%%%%%%%%%%%%%%%%%%%%%%5%%%%%%%%%%%%%%%%%%%%%%%%%%%%%%%%%%%%%%%%%%%
\section{Methods}\label{s:method}

%%%%%%%%%%%%%%%%%%%%%%%%%%%
\subsection{A Quantum-Inspired View of Causal Interaction}\label{ssec:view}

The central proposal of this work is to view interaction not as a special feature term, but as an emergent property of the aggregation rule. We contrast a classical, incoherent rule with a quantum-inspired, coherent rule. In ML terms, we consider two outcome \emph{classes} $y\in\{0,1\}$ (we use ``channel'' and ``class'' interchangeably) and define class-specific complex amplitudes whose squared magnitudes yield probabilities.

\noindent\textbf{Notation.}
We use uppercase $P$ for probabilities and $E$ for unnormalized energies 
(squared amplitude magnitudes); Let $x\in\mathbb{R}^d$ denote the input 
feature vector. For each class $y\in\{0,1\}$, let $\{a_{y,k}(x)\}_{k=1}^{K}\subset\mathbb{C}$ denote $K$ internal amplitude components (class-$y$ "branches"), and define the class amplitude
\[
\psi_y(x)\;=\;\sum_{k=1}^{K} a_{y,k}(x).
\]
Write $\mathrm{Re}(\cdot)$ for the real part and $(\cdot)^{\ast}$ for complex conjugation.

\noindent\textbf{Incoherent vs.\ coherent class energies.}
The incoherent (square-then-sum) class energy removes all cross-terms:
\[
E^{\mathrm{inc}}_{y}(x)\;=\;\sum_{k=1}^{K}\bigl|a_{y,k}(x)\bigr|^{2}.
\]
The coherent (add-then-square) class energy includes interference between components:
\[
E^{\mathrm{coh}}_{y}(x)\;=\;\biggl|\sum_{k=1}^{K} a_{y,k}(x)\biggr|^{2}
\;=\;\sum_{k=1}^{K}\bigl|a_{y,k}(x)\bigr|^{2}
\;+\;\sum_{1\le i<j\le K} 2\,\mathrm{Re}\!\bigl(a_{y,i}(x)\,a_{y,j}(x)^{\ast}\bigr).
\]

\noindent\textbf{From energies to probabilities.}
Binary class probabilities are obtained by normalizing class energies:
\[
P_{\mathrm{inc}}(Y{=}1\mid x)\;=\;\frac{E^{\mathrm{inc}}_{1}(x)}{E^{\mathrm{inc}}_{0}(x)+E^{\mathrm{inc}}_{1}(x)},
\qquad
P_{\mathrm{coh}}(Y{=}1\mid x)\;=\;\frac{E^{\mathrm{coh}}_{1}(x)}{E^{\mathrm{coh}}_{0}(x)+E^{\mathrm{coh}}_{1}(x)}.
\]
Thus the only difference between the two readouts is whether within-class cross-terms are present ($E^{\mathrm{coh}}$) or removed ($E^{\mathrm{inc}}$).

\noindent\textbf{A minimal amplitude-linear parameterization.}
For a practical instantiation we use a class-specific affine complex map
\[
\psi_y(x)\;=\;b_y\;+\;\sum_{j=1}^{d} x_j\,w_{y,j},\qquad w_{y,j},b_y\in\mathbb{C}.
\]
This corresponds to $K{=}1$ (a single amplitude per class).
Then the coherent class energy expands as
\[
E^{\mathrm{coh}}_y(x)\;=\;\bigl|\psi_y(x)\bigr|^{2}
\;=\;|b_y|^{2}
\;+\;2\,\mathrm{Re}\!\Bigl(b_y\,\sum_{j} x_j\,w_{y,j}^{\ast}\Bigr)
\;+\;\sum_{j}|w_{y,j}|^{2}x_j^{2}
\;+\;2\,\mathrm{Re}\!\Bigl(\sum_{i<j}w_{y,i}w_{y,j}^{\ast}\,x_i x_j\Bigr),
\]
whereas the incoherent proxy drops all phase-sensitive cross-terms:
\[
E^{\mathrm{inc}}_y(x)\;\equiv\;|b_y|^{2}\;+\;\sum_{j}|w_{y,j}|^{2}x_j^{2}.
\]
(For one-hot features, $x_j^2=x_j$, so the removal of cross-terms is explicit.)

The two $2\,\mathrm{Re}(\cdot)$ cross-terms absent from $E^{\mathrm{inc}}$ are the interference terms that give rise to interaction effects in coherent aggregation. In the general $d$-dimensional case, these encode all pairwise feature interactions (bias-weight and weight-weight). The next subsection establishes the precise connection to causal interaction for a minimal $2{\times}2$ design.

%%%%%%%%%%%%%%%%%%%%%%%%%%%%%%%%%%%%%%%%%%%%%%%%%%%
\subsection{The Interference--Interaction Identity}\label{ssec:identity}
We connect the classical potential-outcome interaction contrast to the interference cross-term induced by coherent aggregation.

\noindent\textbf{Definition (Interaction contrast).}
Let $(a,b)\in\{0,1\}^2$ denote the two-factor design and $p_{ab}\in[0,1]$ the cell probability
$p_{ab}=P(Y{=}1\mid a,b)$. Define
\[
\Delta_{\mathrm{INT}}\;\equiv\;p_{11}-p_{10}-p_{01}+p_{00}.
\]

\noindent\textbf{Setup.}
We consider a minimal amplitude–linear model with a single complex amplitude whose squared magnitude yields the Bernoulli success probability (one channel):
\[
\psi(a,b)=u_0+a\,u_A+b\,u_B,\qquad u_0,u_A,u_B\in\mathbb{C}.
\]
Under the Born rule, the corresponding cell probability is \(p_{ab}=|\psi(a,b)|^2\).
We choose magnitudes so that \(0\le p_{ab}\le1\) ensures a valid probability; the identity below, which links the interaction contrast to the interference cross-term, is algebraic and does not depend on normalization across outcomes.  This one–channel formulation suffices to establish the identity; the practical binary classifier in Sec.~\ref{ssec:ikc} uses two outcome channels \(y\in\{0,1\}\).

\noindent\textbf{Proposition (Interference--interaction identity).}
With the above definitions,
\[
\Delta_{\mathrm{INT}}
\;=\;2\,\mathrm{Re}\!\bigl(u_Au_B^{\ast}\bigr),
\]
which depends only on the pair \((u_A,u_B)\).

\noindent\textbf{Proof.}
Expand each cell:
\begin{align*}
p_{11}&=|u_0+u_A+u_B|^2
=|u_0|^2+|u_A|^2+|u_B|^2+2\,\mathrm{Re}(u_0u_A^{\ast}+u_0u_B^{\ast}+u_Au_B^{\ast}),\\
p_{10}&=|u_0+u_A|^2=|u_0|^2+|u_A|^2+2\,\mathrm{Re}(u_0u_A^{\ast}),\\
p_{01}&=|u_0+u_B|^2=|u_0|^2+|u_B|^2+2\,\mathrm{Re}(u_0u_B^{\ast}),\\
p_{00}&=|u_0|^2.
\end{align*}
Taking the alternating sum \(p_{11}-p_{10}-p_{01}+p_{00}\) cancels all terms except the cross-term \(2\,\mathrm{Re}(u_Au_B^{\ast})\). \hfill$\square$

\noindent\textbf{Consequences and interpretation.}
Writing \(u_k=r_ke^{i\phi_k}\) (with \(r_k\ge 0\), \(\phi_k\in\mathbb{R}\)) yields
\[
\Delta_{\mathrm{INT}}=2\,r_A r_B \cos(\phi_B-\phi_A),
\]
so (i) the sign of interaction is a geometric property of the relative phase \(\phi_B-\phi_A\) (synergy \(>\!0\), antagonism \(<\!0\)); (ii) the magnitude is bounded by \(2r_Ar_B\); and (iii) the result is invariant to global phase shifts \(u_k\mapsto e^{i\theta}u_k\) and to the background amplitude \(u_0\).
The identity holds algebraically for each cell \((A=a,B=b)\). 

%%%%%%%%%%%%%%%%%%%%%%%%%%%%%%%%%%%%%%%%%%%%%%%%%%%%%%%%
\subsection{The Interference Kernel Classifier (IKC) Model}\label{ssec:ikc}

\noindent\textbf{Instantiation.}
We implement phase-coherent aggregation as a practical binary classifier (IKC).
In the notation of Sec.~\ref{ssec:view}, where \(\psi_y=\sum_{k} a_{y,k}\), we instantiate the class (output channel) amplitude as an affine complex map
\[
\psi_y(x)=w_y^\top x + b_y,\qquad w_y\in\mathbb{C}^d,\; b_y\in\mathbb{C},\;\; y\in\{0,1\},
\]
with \(x\in\mathbb{R}^d\) the preprocessed feature vector. In implementation, we represent \(\psi_y(x)=\alpha_y(x)+\mathrm{i}\,\beta_y(x)\) using two real-valued affine maps per class; the relative phases among coefficients determine the sign and magnitude of the within-class interference terms.

\noindent\textbf{Born-aggregated probability.}
The IKC predictive distribution is given by the coherent probability 
$P_{\mathrm{IKC}}(Y{=}1\mid x) = P_{\mathrm{coh}}(Y{=}1\mid x)$ 
defined above, where $E^{\mathrm{coh}}_y(x) = |\psi_y(x)|^2 
= \alpha_y(x)^2+\beta_y(x)^2$.
This map is invariant to a common nonzero complex rescaling,
i.e., $\psi_y \mapsto c\,\psi_y$ for any $c\in\mathbb{C}\setminus\{0\}$ 
leaves $P_{\mathrm{IKC}}$ unchanged.

\noindent\textbf{Training objective and regularization.}
Parameters \(\{w_y,b_y\}\) are learned by minimizing the negative log-likelihood (NLL) on the training split using the Born probabilities. For numerical stability we clip probabilities to $[10^{-7},\,1-10^{-7}]$. We apply $\ell_2$ weight decay (i.e., an $L_2$ regularization term on parameter magnitudes) to the real-valued parameterization $(\alpha,\beta)$, which helps stabilize the global scale non-identifiability noted above.  Optimization uses first-order methods (e.g., Adam~\cite{kingma:2015:adam}); the optimizer and its hyperparameters are selected by budget-matched randomized search.

\noindent\textbf{Temperature scaling.}
We apply post-hoc TS as specified in the common protocol (Sec.~\ref{s:experiments}): a single temperature is fit on the calibration split and used with the safety-switch policy described there.

\noindent\textbf{Incoherent toggle for ablation (Sec.~\ref{ssec:exp4}).}
Holding the learned parameters fixed, we isolate the aggregation rule by switching the readout to the incoherent proxy $P_{\mathrm{inc}}(Y{=}1\mid x)$ defined in Sec.~\ref{ssec:view}, which removes all phase-sensitive cross-terms.  This toggle changes only the aggregation rule (add-then-square $\rightarrow$ square-then-sum) and leaves architecture, parameters, and splits unchanged.

\noindent\textbf{Diagnostics.}
Alongside standard metrics (NLL, Brier, and ECE), we report two coherence diagnostics.  Let $P_{\mathrm{coh}}(\cdot\mid x)$ and $P_{\mathrm{inc}}(\cdot\mid x)$ denote the binary class probabilities under coherent and incoherent aggregation, respectively, using identical learned parameters (the incoherent proxy removes within-class cross-terms only).

\emph{Coherent Gain} quantifies the average per-instance log-likelihood improvement:
\[
G_{\mathrm{coh}}
\;=\;
\frac{1}{n}\sum_{i=1}^{n}
\Bigl[\log P_{\mathrm{coh}}(y_i\mid x_i)\;-\;\log P_{\mathrm{inc}}(y_i\mid x_i)\Bigr].
\]

\emph{Interference Information} measures the distributional shift induced by coherence via the average KL divergence:
\[
\mathcal{J}_{\mathrm{int}}
\;=\;
\frac{1}{n}\sum_{i=1}^{n}
\mathrm{KL}\!\bigl(P_{\mathrm{coh}}(\cdot\mid x_i)\,\|\,P_{\mathrm{inc}}(\cdot\mid x_i)\bigr),
\quad
\text{with }\mathrm{KL}(p\|q)=\sum_{y\in\{0,1\}} p_y\log\frac{p_y}{q_y}.
\]

Both diagnostics are computed from raw (pre–TS) probabilities and quantify the \emph{local} (within-representation) benefit of coherent aggregation: holding learned parameters fixed, $G_{\mathrm{coh}} > 0$ indicates that phase-sensitive composition improves per-instance likelihood relative to an incoherent proxy, and $\mathcal{J}_{\mathrm{int}} > 0$ measures the distributional shift induced by the interference cross-term. These diagnostics isolate the aggregation rule's contribution from representational capacity, complementing global performance metrics (NLL, Brier, and ECE) that conflate both.  Note that $\mathcal{J}_{\mathrm{int}}\ge 0$ by the non-negativity of KL divergence. 

%%%%%%%%%%%%%%%%%%%%%%%%%%%%%%%%%%%%%%%%%%%%%%%%%%%%%%%%%%%%%%%%%
\section{Experimental Setup}\label{s:experiments}

We evaluate the IKC in five experiments that progress from theory to practice:
\textbf{(E1)} identity validation on a minimal \(2{\times}2\) synthetic design;
\textbf{(E2)} a high-interaction XOR task compared against strong baselines;
\textbf{(E3)} robustness to training-only label noise with a TS safety switch;
\textbf{(E4)} an ablation that toggles coherent vs.\ incoherent aggregation while keeping learned parameters fixed; and
\textbf{(E5)} real-world benchmarks on \emph{Adult} and \emph{Bank Marketing} with calibrated evaluation.
Detailed protocols for each experiment follow.

\noindent\textbf{Common protocol.}
Unless otherwise noted, all experiments follow this protocol:
\begin{itemize}
  \item \textbf{Splits:} For each dataset and seed, we use fixed training, validation, calibration, and test splits (61.25\%, 15\%, 8.75\%, and 15\%).
  \item \textbf{Hyperparameter optimization:} We run budget-matched randomized HPO for each model and select the configuration with the lowest validation NLL.
  \item \textbf{Retraining:} The selected configuration is retrained on the union of the training and validation splits before final evaluation.
  \item \textbf{Temperature scaling:} A single temperature is fit on the calibration split and applied to test predictions only if it lowers calibration NLL (safety switch); otherwise, raw probabilities are retained.
  \item \textbf{Metrics:} We report NLL, Brier score, ECE (15 equal-width bins), and accuracy on the test split.
  \item \textbf{Statistical analysis:} Across seeds, we aggregate with means and two-sided bootstrap 95\% CIs (5,000 replicates), and compute exact paired sign-flip \(p\)-values.
\end{itemize}

\noindent\textbf{Deviations from common protocol.}
(E1) Follows its own simulation protocol (not using dataset splits nor TS).  Table~\ref{tab:exp-protocol-summary} summarizes the procedural differences across experiments.

\begin{table}[t]
\centering
\caption{Summary of selection and evaluation differences across experiments. E2 uses $n_{\text{seeds}}=10$ (smaller sample for rapid iteration on synthetic data); E3, E4, and E5 use $n_{\text{seeds}}=20$.}
\label{tab:exp-protocol-summary}
\footnotesize
\begin{adjustbox}{max width=\textwidth}
\begin{tabular}{@{}>{\centering\arraybackslash}m{0.13\textwidth} m{0.31\textwidth} m{0.28\textwidth} m{0.26\textwidth}@{}}
\toprule
\textbf{Experiment} & \textbf{Baseline selection criterion} & \textbf{Retraining on training and validation} & \textbf{TS policy} \\
\midrule
\makecell[c]{\textbf{E2}\\(XOR)} &
Best classical baseline chosen by test NLL on the fixed test split (within seed) &
\textbf{No} (comparison uses validation-tuned models without retraining on training and validation) &
\textbf{Always applied} (no safety switch; TS fit on calibration and applied to test) \\
\addlinespace[0.35em]
\makecell[c]{\textbf{E3}\\(Robustness)} &
Best classical baseline chosen by validation NLL (within seed and noise level) &
\textbf{Yes} (selected configuration retrained on training and validation before test) &
\textbf{Safety switch} (apply TS to test only if it lowers calibration NLL) \\
\addlinespace[0.35em]
\makecell[c]{\textbf{E4}\\(Ablation)} &
N/A (coherent vs.\ incoherent toggle; no baseline comparison) &
\textbf{Yes} (default protocol) &
\textbf{Safety switch} independently for each mode \\
\addlinespace[0.35em]
\makecell[c]{\textbf{E5}\\(Real-world)} &
Anchor (best) baseline chosen by test NLL on the fixed test split (within seed) &
\textbf{Yes} (default protocol: retraining on training and validation for final evaluation) &
\textbf{Safety switch} (apply TS to test only if it lowers calibration NLL) \\
\bottomrule
\end{tabular}
\end{adjustbox}
\end{table}

%%%%%%%%%%%%%%%%%%%%%%%%
\subsection{Protocol for Experiment 1 (E1): Phase Sweep Validation}\label{ssec:exp1}

\noindent\textbf{Objective.}
Provide a direct, quantitative validation of the identity between the potential-outcome interaction contrast and the interference cross-term.

\noindent\textbf{Theoretical model.}
We use the minimal amplitude-linear model (Sec.~\ref{ssec:identity}), where \(\Delta_{\mathrm{INT}} = 2\,r_A r_B\,\cos(\Delta\phi)\) with \(\Delta\phi\equiv\phi_B-\phi_A\) the relative phase.

\noindent\textbf{Simulation procedure.}
Fix \((r_0,r_A,r_B)=(0.20,0.35,0.35)\) and sweep the relative phase \(\Delta\phi\) over 121 points in \([-\pi,\pi]\).  For each \(\Delta\phi\), simulate a balanced \(2{\times}2\) design with \(n=2000\) Bernoulli trials per cell, estimate \(\widehat{p}_{ab}\), and compute \(\widehat{\Delta}_{\mathrm{INT}}\).
Without loss of generality set \(\phi_A=0\) and realize the sweep by \(\phi_B=\Delta\phi\).
(The magnitudes keep \(p_{ab}\) away from \(0/1\) and match the theoretical range \(2r_A r_B=0.245\).)

\noindent\textbf{Statistical analysis.}
Two-sided 95\% bootstrap CIs are formed with 5{,}000 replicates (seed 108).

%%%%%%%%%%%%%%%%%%%%%%%%
\subsection{Protocol for Experiment 2 (E2): Comparison with Strong Baselines}\label{ssec:exp2}

\noindent\textbf{Objective.}
Compare the coherent (Born-aggregating) model against strong classical baselines on a canonical high-interaction task (XOR) to assess performance when the signal resides purely in interaction.

\noindent\textbf{Data-generating process.}
Inputs are binary and independent, $X_1,X_2 \sim \mathrm{Bernoulli}(1/2)$, with the label $Y=X_1 \oplus X_2$.  We generate $N=20{,}000$ samples.  A single, fixed test split of $3{,}000$ samples (15\%) is created using the first random seed and is reused for all models and seeds to enable paired comparisons.  The remaining $17{,}000$ samples are, for each seed, partitioned into training (61.25\%), validation (15\%), and calibration (8.75\%).

\noindent\textbf{Models and baselines.}
We compare IKC with the following baselines:
\begin{itemize}
  \item \textbf{Logistic Regression with an Interaction Term.} Logistic regression on degree-2 polynomial features including the product $X_1X_2$.
  \item \textbf{Quadratic Logistic Regression.} Logistic regression using the features $X_1$, $X_2$, and their squares $X_1^2$, $X_2^2$. For binary inputs (where $X_i^2=X_i$), this reduces to a main-effects-only model without interaction, serving as a deliberate mis-specification baseline.
  \item \textbf{XGBoost.} A strong tree-based baseline.
\end{itemize}

\noindent\textbf{Training and hyperparameter search.}
For each of $n_{\text{seeds}}=10$ seeds, we run randomized HPO with a fixed budget of twenty trials per model under the same search space, early-stopping policy, and seeding policy.  The configuration with the lowest \emph{validation} NLL is selected for each model (intra-model selection).  For paired comparisons, we then identify the classical baseline with the lowest \emph{test} NLL on the fixed test split (inter-model anchor selection). This design provides a conservative comparison against the empirically strongest baseline on the test set, effectively tilting the comparison against IKC, which may introduce upward bias in baseline performance due to post-hoc anchor selection.  No model is retrained or further tuned using test data.

%%%%%%%%%%%%%%%%%%%%%%%%%%%%%%%%%%%%%
\subsection{Protocol for Experiment 3 (E3): Robustness Evaluation}\label{ssec:exp3}

\noindent\textbf{Objective.}
Evaluate robustness under controlled label corruption, focusing on how performance degrades as noise increases and whether IKC's advantages, if any, persist.

\noindent\textbf{Datasets and preprocessing.}
We use \emph{Adult} (Census Income; UCI repository) to predict whether annual income exceeds \$50K and \emph{Bank Marketing} (UCI) to predict whether a client subscribes to a term deposit~\cite{adult_2,bank_marketing_222}. 
Categorical features undergo mode imputation and one-hot encoding; 
numerical features undergo median imputation and standardization.
For \emph{Bank}, we exclude \texttt{duration} because it is observed only after the call and leaks target information, and we treat both ``?'' and ``unknown'' as missing~\cite{bank_marketing_222}.
Table~\ref{tab:data_prep} summarizes datasets and preprocessing.
For each seed, we create fixed splits; within a seed, the same test split is reused across all noise levels to enable paired comparisons.

\begin{table}[t]
\centering
\caption{Datasets and preprocessing used in E3 and E5.}
\label{tab:data_prep}
\begin{adjustbox}{max width=\textwidth}
\begin{tabular}{@{}l p{0.22\textwidth} p{0.60\textwidth}@{}}
\toprule
\textbf{Dataset} & \textbf{Target} & \textbf{Preprocessing} \\
\midrule
Adult & \texttt{income} (>\,50K) &
\begin{minipage}[t]{\linewidth}
\begin{itemize}[leftmargin=*,labelsep=0.4em,topsep=0pt,itemsep=0pt,parsep=0pt,partopsep=0pt]
  \item Categorical: impute (mode) $\to$ OneHotEncoder (\texttt{handle\_unknown=ignore})
  \item Numeric: impute (median) $\to$ standardize
\end{itemize}
\end{minipage}
\\[0.25em]
Bank & \texttt{y} (= yes) &
\begin{minipage}[t]{\linewidth}
\begin{itemize}[leftmargin=*,labelsep=0.4em,topsep=0pt,itemsep=0pt,parsep=0pt,partopsep=0pt]
  \item Same as Adult
  \item Drop \texttt{duration}
  \item Treat ``?'' and ``unknown'' as missing
\end{itemize}
\end{minipage}
\\
\bottomrule
\end{tabular}
\end{adjustbox}
\end{table}

\noindent\textbf{Noise injection.}
Corruption is applied only to the training split; validation, calibration, and test remain clean to avoid leakage.  We inject symmetric label noise by flipping the training label independently with probability $p \in \{0,\,0.05,\,0.10,\,0.15,\,0.20,\,0.25,\,0.30\}$.

\noindent\textbf{Models.}
Same as E2.

\noindent\textbf{Deviations from common protocol.}
The best baseline is selected by validation NLL within each (noise level, seed) pair. We use $n_{\text{seeds}}=20$.

%%%%%%%%%%%%%%%
\subsection{Protocol for Experiment 4 (E4): Ablation on Coherent vs.\ Incoherent Aggregation}\label{ssec:exp4}

\noindent\textbf{Objective.}
Isolate the contribution of the aggregation rule by evaluating a single trained IKC model in two modes that differ only at the final aggregation step: (i) coherent (add amplitudes then square; cross terms present) and (ii) an incoherent proxy (square then sum; all cross terms removed). Holding all learned parameters fixed, we test whether the interference cross term improves likelihood and calibration.

\noindent\textbf{Datasets and preprocessing.}
We use \emph{Adult} and \emph{Bank} with the E3 preprocessing pipeline. 

\noindent\textbf{Ablation design.}
For each trained IKC model (parameterized by \(\{w_y,b_y\}_{y\in\{0,1\}}\) with \(\psi_y(x)=w_y^\top x+b_y\)), we form two probability distributions using identical parameters:

\textit{Coherent mode:} Uses $P_{\mathrm{coh}}(Y{=}1\mid x)$ with $E^{\mathrm{coh}}_y(x) = |\psi_y(x)|^2$ (Born aggregation retains phase-sensitive cross-terms).

\textit{Incoherent mode:} Uses $P_{\mathrm{inc}}(Y{=}1\mid x)$ with $E^{\mathrm{inc}}_y(x) = |b_y|^2+\sum_{j}|w_{y,j}|^2\,x_j^2$ (cross-terms removed, including bias-feature and feature-feature interactions; for one-hot features, $x_j^2=x_j$).

Both modes use the same Born normalization and parameters (see Sec.~\ref{ssec:ikc} for derivations).

\noindent\textbf{Deviations from common protocol.}
TS is applied independently for each mode using the safety switch. Diagnostics \(\{G_{\mathrm{coh}},\,\mathcal{J}_{\mathrm{int}}\}\) are computed from raw (pre-TS) probabilities. Paired differences $\Delta_{\mathrm{coh-inc}} \equiv m_{\mathrm{coh}} - m_{\mathrm{inc}}$ (where $m \in \{\text{NLL},\text{Brier},\text{ECE}\}$) are analyzed. We use $n_{\text{seeds}}=20$.

%%%%%%%%%%%%%%%%%%%%%%%%%%%
\subsection{Protocol for Experiment 5 (E5): Real-World Application}\label{ssec:exp5}

\noindent\textbf{Objective.}
Assess whether the coherent (Born-aggregating) model provides practical benefit on real tabular problems, using two standard benchmarks (\emph{Adult} and \emph{Bank}). We emphasize fair training budgets, calibrated evaluation, and paired statistical comparisons against strong baselines.

\noindent\textbf{Datasets and preprocessing.}
We use \emph{Adult} and \emph{Bank} with the preprocessing pipeline from E3 (see Table~\ref{tab:data_prep} for details).

\noindent\textbf{Models and hyperparameter search.}
We use the same models as in E2 (Sec.~\ref{ssec:exp2}): IKC, logistic regression with pairwise interactions, quadratic logistic regression, and XGBoost.  Within each seed and dataset, every model receives the same randomized hyperparameter search budget (25 trials). The configuration with the lowest validation NLL is selected, and the model is retrained on the union of the training and validation splits before evaluation.

\noindent\textbf{Metrics.}
Per seed, we anchor to the classical baseline with the lowest test NLL--a conservative choice against IKC--and compute paired differences \((\text{IKC}-\text{Best-Baseline})\) for NLL, Brier, and ECE. For interpretability, we additionally compute \{\(G_{\mathrm{coh}},\mathcal{J}_{\mathrm{int}}\}\) from the IKC model's raw (pre-TS) probabilities.

%%%%%%%%%%%%%%%%%%%%%%%%%%%%%%%%%%%%%%%%%%%%%%%%%%%%%%%%%%%%%%%%
\section{Experimental Results}\label{s:results}

All experiments were run in Python~3 with \texttt{PyTorch}~\cite{pytorch}, \texttt{NumPy}~\cite{harris:2020:numpy}, \texttt{SciPy}~\cite{virtanen:2020:scipy}, and \texttt{Matplotlib}~\cite{hunter:2007:matplotlib} for numerical routines and visualization.
Classical baselines used \texttt{scikit{-}learn}~\cite{scikit-learn} and XGBoost~\cite{chen:2016:xgboost}.
Our IKC implementation follows the methods in Sec.~\ref{ssec:ikc}; TS and sign-flip tests use the same stack.

\subsection{E1: Phase Sweep Validation of the Interference--Interaction Identity}

\begin{figure}[htbp!]
  \centering
  \includegraphics[width=0.85\linewidth]{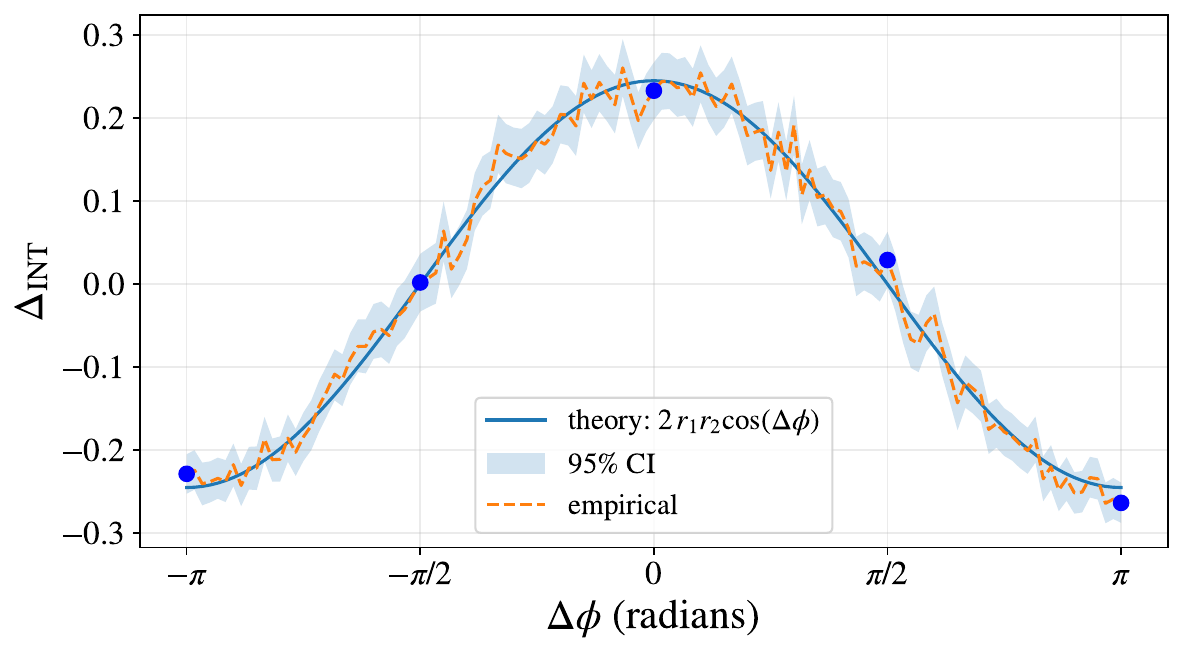}
  \caption{\textbf{E1: Phase sweep validates the interference–interaction identity.}
  Interaction contrast \(\Delta_{\mathrm{INT}}\) as a function of the relative phase \(\Delta\phi\).
  The plot overlays the theoretical law \(2\,r_A r_B \cos(\Delta\phi)\) from the amplitude–linear model with \((r_0,r_A,r_B)=(0.20,0.35,0.35)\) and empirical plug-in estimates \(\widehat{\Delta}_{\mathrm{INT}}\) with two-sided 95\% bootstrap confidence bands (balanced \(2\times 2\) design; \(n{=}2000\) per cell; 121 phase points; 5{,}000 bootstrap replicates; a fixed random seed).}
  \label{fig:e1_sweep}
\end{figure}

We validate the identity from Sec.~\ref{ssec:identity} by sweeping the relative phase \(\Delta\phi\) over 121 points (see Sec.~\ref{ssec:exp1} for parameters).  The theoretical prediction is \(\Delta_{\mathrm{INT}} = 2\,r_A r_B\,\cos(\Delta\phi) = 0.245\,\cos(\Delta\phi)\).  Figure~\ref{fig:e1_sweep} shows that empirical estimates \(\widehat{\Delta}_{\mathrm{INT}}\) closely track the theoretical curve. Representative checkpoints:
\begin{itemize}
\item \(\Delta\phi\approx -\pi\): theory \(-0.245\), 
  observed \(-0.229\,[-0.253,-0.205]\).
\item \(\Delta\phi\approx -\pi/2\): theory \(0.000\), 
  observed \(+0.002\,[-0.033,+0.037]\).
\item \(\Delta\phi\approx 0\): theory \(+0.245\), 
  observed \(+0.233\,[+0.198,+0.267]\).
\item \(\Delta\phi\approx +\pi/2\): theory \(0.000\), 
  observed \(+0.029\,[-0.004,+0.063]\).
\item \(\Delta\phi\approx +\pi\): theory \(-0.245\), 
  observed \(-0.264\,[-0.288,-0.239]\).
\end{itemize}
At neutral points (\(\pm\pi/2\)), confidence intervals include zero; 
at extremes (\(0,\pm\pi\)), they exclude zero with the correct sign.

%%%%%%%%%% 
\subsection{E2: Comparison with Strong Baselines}

\noindent\textbf{Main results.}
Paired, budget-matched comparisons \((\text{IKC}-\text{Best-Baseline})\) 
show that IKC significantly outperforms on XOR (Table~\ref{tab:e2-main}; 
\(n_{\text{seeds}}{=}10\)):
\(\Delta\text{NLL}=-0.159\pm0.134\) (exact \(p=0.016\)),
\(\Delta\text{Brier}=-0.062\pm0.057\),
and \(\Delta\text{ECE}=-0.034\pm0.023\) (exact \(p{\approx}0.121\)).
IKC yields clear likelihood and Brier improvements; the ECE reduction 
is consistent in magnitude but not statistically significant under 
the sign-flip test.
Because the anchor baseline is chosen post-hoc from \(K{=}3\) candidates, 
the reported \(p\)-value may be optimistic. Even under a conservative 
Bonferroni correction (threshold: 0.05/3 $\approx$ 0.017), the result remains 
significant (\(p{=}0.016\)), though this adjustment may be overly stringent 
given the paired design on a fixed test split.

\noindent\textbf{Interpretation.}
XOR is a pure-interaction task: main effects are zero, and only the interaction term is informative.  IKC's coherent aggregation captures this via the phase-sensitive cross-term in the amplitude-linear model, whereas baselines that omit the explicit interaction (e.g., \emph{Quadratic}) are mis-specified; even when the interaction feature is present (logistic regression with an interaction term), IKC's Born aggregation confers a likelihood advantage under the same calibration protocol.

\noindent\textbf{Interference diagnostics.}
From IKC's raw probabilities, both \(G_{\mathrm{coh}}\) and \(\mathcal{J}_{\mathrm{int}}\) are strongly positive with tight CIs: \(G_{\mathrm{coh}}=0.443\pm0.062\) and \(\mathcal{J}_{\mathrm{int}}=0.456\pm0.061\) (exact \(p<0.01\)), quantifying a per-instance log-likelihood edge over the incoherent proxy and a nonzero KL gap induced by the cross-term.

\noindent\textbf{Takeaway.}
On a canonical high-interaction task, coherent (Born) aggregation delivers a statistically significant likelihood and Brier improvements over strong classical baselines under matched budgets and identical calibration.

\begin{table}[t]
\centering
\caption{E2 (XOR): mean paired differences $(\text{IKC}-\text{Best-Baseline})$ across seeds (lower is better). Values are mean $\pm$ 95\% CI half-width; exact $p$ from a paired sign-flip test on NLL. TS is fit on the calibration split and applied to test predictions.}
\label{tab:e2-main}
\setlength{\tabcolsep}{6pt}
\footnotesize
\begin{adjustbox}{max width=\linewidth}
\begin{tabular}{@{}lcccc@{}}
\toprule
\(n_{\text{seeds}}\) & \(\Delta\)NLL & \(\Delta\)Brier & \(\Delta\)ECE & exact \(p\) (NLL) \\
\midrule
10 & \(-0.159\pm 0.134\) & \(-0.062\pm 0.057\) & \(-0.034\pm 0.023\) & \(0.016\) \\
\bottomrule
\end{tabular}
\end{adjustbox}
\end{table}

%%%%%%%%%%%%%%%%%%%%%%%%%%%%%%%%%%%%%%%%%%%%%%%%%%%%%%%%%%%%%% 
\subsection{E3: Robustness to Label Noise}

\begin{figure}[htbp!]
  \centering
  \includegraphics[width=0.99\linewidth]{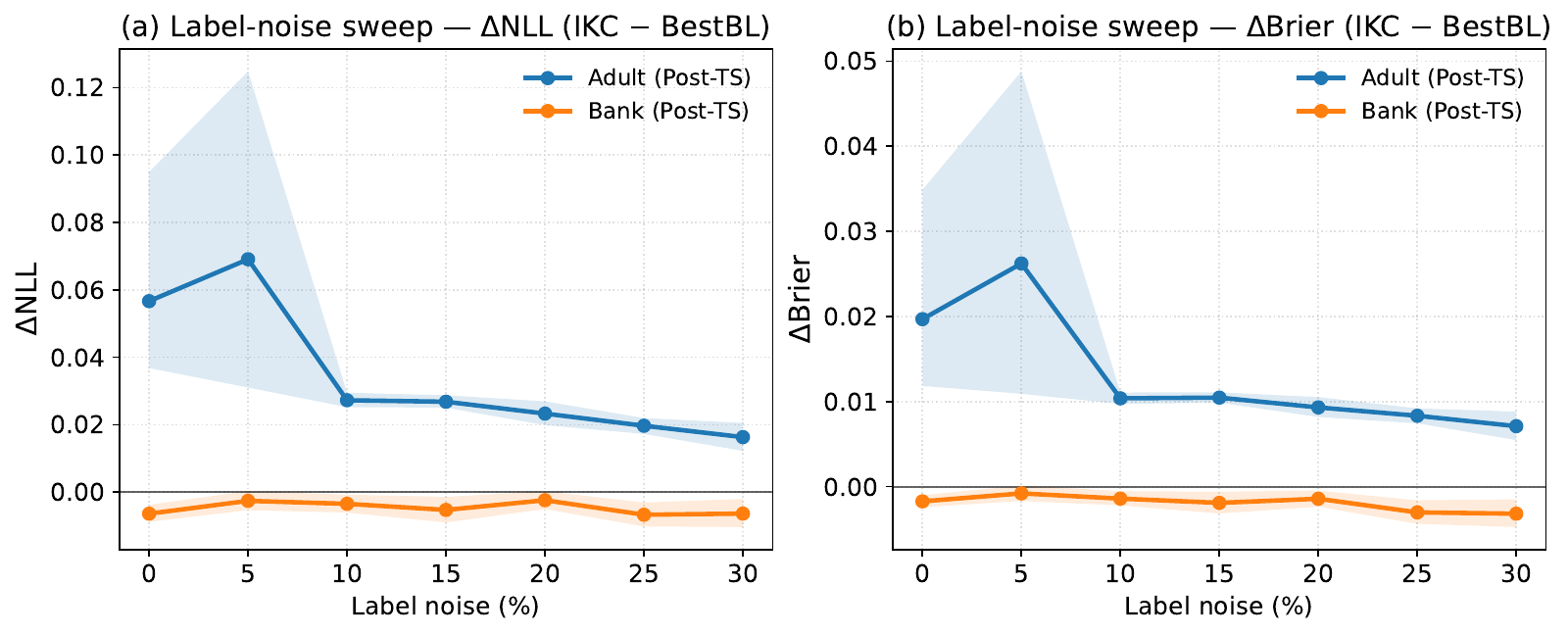}
  \caption{\textbf{E3: Robustness to training label noise.}
  Paired deltas (\(\Delta=\) IKC \(-\) Best-Baseline; lower is better) versus label-noise level (\%) with two-sided 95\% bootstrap CIs, using a fixed test split per seed.
  \textbf{(a)} \(\Delta\)NLL; \textbf{(b)} \(\Delta\)Brier.
  The best baseline is chosen per seed by validation NLL on the same split and retrained on the training and validation data.
  TS uses a safety switch (fit on the calibration split; applied to the test split only if it lowers calibration NLL)}
  \label{fig:e3-label}
\end{figure}

Figure~\ref{fig:e3-label} shows the paired deltas after TS versus label-noise level for both datasets: panel (a) displays \(\Delta\)NLL and panel (b) \(\Delta\)Brier. Solid lines denote means; shaded bands are two-sided 95\% bootstrap CIs (fixed test split per seed).

\noindent\textbf{Adult (Table~\ref{tab:e3-adult-robust}).}
IKC underperforms the best baseline on likelihood and Brier across all noise rates. For example, \(\Delta\text{NLL}=+0.057 \pm 0.029\) at 0\% noise (\(p=3.8{\times}10^{-6}\)) and remains positive under heavy noise (\(+0.016 \pm 0.004\) at 30\%, \(p=7.6{\times}10^{-6}\)). Brier shows the same pattern (\(+0.020 \pm 0.012\) at 0\%; \(+0.007 \pm 0.002\) at 30\%). ECE gaps are small and often favor IKC at higher noise (e.g., \(-0.006 \pm 0.003\) at 25\%), but these calibration gains do not offset the likelihood disadvantage.

\noindent\textbf{Bank (Table~\ref{tab:e3-bank-robust}).}
On \emph{Bank}, IKC generally outperforms the best baseline on NLL and Brier, including the clean case: \(\Delta\text{NLL}=-0.006 \pm 0.003\) at 0\% noise (\(p=8.9{\times}10^{-4}\)) with consistent Brier gains (\(-0.002 \pm 0.001\)). Improvements remain statistically significant at most noise levels (e.g., 10\%: \(-0.004 \pm 0.003\), \(p=0.047\); 15\%: \(-0.005 \pm 0.004\), \(p=0.034\); 25\%: \(-0.007 \pm 0.004\), \(p=0.006\); 30\%: \(-0.006 \pm 0.004\), \(p=0.022\)), while 5\% and 20\% are not significant. ECE effects are mixed (e.g., modest degradation at 20\%, \(+0.005 \pm 0.004\)), but the likelihood and Brier gains dominate.

\noindent\textbf{Takeaway.}
Under training-only label noise, IKC is less competitive on \emph{Adult} yet delivers consistent likelihood and Brier improvements on \emph{Bank} under matched budgets and identical calibration. The safety-switch calibration ensures that reported trends are not artifacts of miscalibration.

\begin{table}[t]
\centering
\footnotesize
\caption{\textbf{E3: robustness on Adult (label noise on training only).} Paired, budget-matched deltas ($\text{IKC}-\text{Best-Baseline}$; lower is better). All models use TS with a safety switch (fit on the clean calibration split; applied to the test split only if it lowers calibration NLL). Values are mean \(\pm\) 95\% CI half-width across \(n_\text{seeds}{=}20\); exact \(p\) from a paired sign-flip test on NLL}
\label{tab:e3-adult-robust}
\begin{adjustbox}{max width=\textwidth}
\begin{tabular}{lcccc}
\toprule
Label noise (\%) & \(\Delta\)NLL & \(\Delta\)Brier & \(\Delta\)ECE & exact \(p\) (NLL) \\
\midrule
 0  & \(0.057 \pm 0.029\) & \(0.020 \pm 0.012\) & \(0.016 \pm 0.020\) & \(3.8\times 10^{-6}\) \\
 5  & \(0.069 \pm 0.047\) & \(0.026 \pm 0.019\) & \(0.023 \pm 0.028\) & \(3.8\times 10^{-6}\) \\
10  & \(0.027 \pm 0.002\) & \(0.010 \pm 0.001\) & \(-0.003 \pm 0.002\) & \(3.8\times 10^{-6}\) \\
15  & \(0.027 \pm 0.002\) & \(0.011 \pm 0.001\) & \(-0.004 \pm 0.004\) & \(3.8\times 10^{-6}\) \\
20  & \(0.023 \pm 0.004\) & \(0.009 \pm 0.001\) & \(-0.003 \pm 0.004\) & \(3.8\times 10^{-6}\) \\
25  & \(0.020 \pm 0.002\) & \(0.008 \pm 0.001\) & \(-0.006 \pm 0.003\) & \(3.8\times 10^{-6}\) \\
30  & \(0.016 \pm 0.004\) & \(0.007 \pm 0.002\) & \(-0.005 \pm 0.004\) & \(7.6\times 10^{-6}\) \\
\bottomrule
\end{tabular}
\end{adjustbox}
\end{table}

\begin{table}[t]
\centering
\footnotesize
\caption{\textbf{E3: robustness on Bank (label noise on training only).} Same setup as Table~\ref{tab:e3-adult-robust}. Values are mean \(\pm\) 95\% CI half-width across \(n_\text{seeds}{=}20\); exact \(p\) from a paired sign-flip test on NLL}
\label{tab:e3-bank-robust}
\begin{adjustbox}{max width=\textwidth}
\begin{tabular}{lcccc}
\toprule
Label noise (\%) & \(\Delta\)NLL & \(\Delta\)Brier & \(\Delta\)ECE & exact \(p\) (NLL) \\
\midrule
 0  & \(-0.006 \pm 0.003\) & \(-0.002 \pm 0.001\) & \(-0.003 \pm 0.002\) & \(8.9\times 10^{-4}\) \\
 5  & \(-0.003 \pm 0.003\) & \(-0.001 \pm 0.001\) & \(0.001 \pm 0.003\)  & \(0.200\) \\
10  & \(-0.004 \pm 0.003\) & \(-0.001 \pm 0.001\) & \(0.003 \pm 0.003\)  & \(0.047\) \\
15  & \(-0.005 \pm 0.004\) & \(-0.002 \pm 0.001\) & \(-0.001 \pm 0.003\) & \(0.034\) \\
20  & \(-0.002 \pm 0.003\) & \(-0.001 \pm 0.001\) & \(0.005 \pm 0.004\)  & \(0.183\) \\
25  & \(-0.007 \pm 0.004\) & \(-0.003 \pm 0.001\) & \(0.001 \pm 0.004\)  & \(0.006\) \\
30  & \(-0.006 \pm 0.004\) & \(-0.003 \pm 0.002\) & \(0.003 \pm 0.004\)  & \(0.022\) \\
\bottomrule
\end{tabular}
\end{adjustbox}
\end{table}

%%%%%%%%%%%%%%%%%%%%%%%%%%%%%%%%%%%%%%%%%%%%%%%
\subsection{E4: Ablation -- Coherent vs.\ Incoherent Aggregation}

We isolate the aggregation effect by evaluating the same trained parameters in coherent and incoherent modes, with TS applied independently to each (see Sec.~\ref{ssec:exp4}).

\noindent\textbf{Main results (Table~\ref{tab:e4_combined}).}
Holding learned parameters fixed, coherent aggregation consistently improves likelihood, Brier score, and calibration on both datasets. On \emph{Adult}, \(\Delta\text{NLL}=-0.102\) (exact \(p{=}3.05{\times}10^{-5}\)); on \emph{Bank}, \(\Delta\text{NLL}=-0.119\) (exact \(p{=}3.81{\times}10^{-6}\)). All metrics favor coherent aggregation with tight confidence intervals.

\noindent\textbf{Interference diagnostics (Table~\ref{tab:e4_combined}).}
Coherent Gain (\(G_{\mathrm{coh}}\)) and Interference Information 
(\(\mathcal{J}_{\mathrm{int}}\)) are strictly positive on both datasets, 
confirming a systematic per-instance log-likelihood advantage of coherent 
over incoherent aggregation.

\noindent\textbf{Takeaway.}
Holding the model and data fixed, flipping only the aggregation rule yields consistent improvements to likelihood, squared error, and calibration across both datasets, isolating the benefit of coherent (Born) aggregation.

\begin{table}[t]
\centering
\caption{E4 (Ablation): Coherent versus incoherent aggregation (same learned parameters; aggregation rule toggled).
Values are mean \([95\%\,\mathrm{CI}]\) across seeds; \(\Delta=\text{IKC}_{\text{coh}}-\text{IKC}_{\text{inc}}\) (lower is better).
Interference metrics (\(G_{\text{coh}}\) and \(\mathcal{J}_{\text{int}}\)) are computed from raw (pre–TS) probabilities.
Exact \(p\) from a paired sign-flip test on NLL; \(n_\text{seeds}{=}20\) per dataset.}
\label{tab:e4_combined}
\footnotesize
\renewcommand{\arraystretch}{1.15}
\setlength{\tabcolsep}{6pt}
\begin{tabular}{@{}lcc@{}}
\toprule
\textbf{Metric} & \textbf{Adult} & \textbf{Bank} \\
\midrule
\(\Delta\)NLL &
\(-0.102\;[{-}0.119,\,{-}0.081]\) &
\(-0.119\;[{-}0.136,\,{-}0.104]\) \\
\(\Delta\)Brier &
\(-0.033\;[{-}0.038,\,{-}0.026]\) &
\(-0.037\;[{-}0.044,\,{-}0.031]\) \\
\(\Delta\)ECE &
\(-0.059\;[{-}0.069,\,{-}0.047]\) &
\(-0.134\;[{-}0.153,\,{-}0.118]\) \\
exact \(p\) (NLL) &
\(3.05\times 10^{-5}\) &
\(3.81\times 10^{-6}\) \\
\(G_{\text{coh}}\) &
\(0.120\;[0.116,\,0.125]\) &
\(0.119\;[0.104,\,0.136]\) \\
\(\mathcal{J}_{\text{int}}\) &
\(0.120\;[0.114,\,0.125]\) &
\(0.124\;[0.109,\,0.141]\) \\
\bottomrule
\end{tabular}
\end{table}

%%%%%%%%%%%%%%%%%%%%%%%%%%%%%%%%%%%%%%%%%%%%%%%
\subsection{E5: Real-World Application}

\noindent\textbf{Main results.}
Unlike E3—which injected symmetric label noise only into the training split—E5 evaluates clean, real-world test sets under the TS safety switch. In this setting the best classical baseline is, on average, superior; consequently, paired deltas (IKC \(-\) Best-Baseline) are positive for NLL, Brier, and ECE on both \emph{Adult} and \emph{Bank}. However, E4 shows that the aggregation rule itself is beneficial: holding learned parameters fixed, coherent aggregation improves NLL, Brier, and ECE over the incoherent proxy. This indicates that the E5 gaps reflect representational capacity rather than the aggregation rule.  While E3 shows advantages under training-label noise, E5 on clean data shows a small disadvantage, suggesting noise tolerance rather than raw capacity as the main driver.

\begin{figure}
  \centering
  \includegraphics[width=0.91\linewidth]{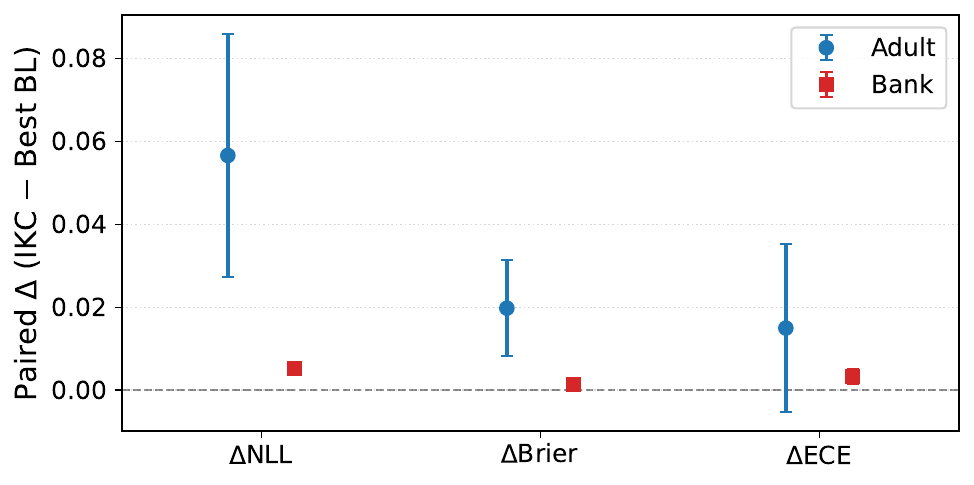}
  \caption{\textbf{E5: Adult and Bank paired test-set deltas.}
  Paired, budget-matched differences \(\Delta=\) IKC \(-\) Best-Baseline (lower is better) for NLL, Brier, and ECE; mean \(\pm\) \(95\%\) CI over \(n_\text{seeds}{=}20\). Best baseline is chosen per seed; all models use TS with a safety switch.}
  \label{fig:e5_deltas_single}
\end{figure}

Figure~\ref{fig:e5_deltas_single} summarizes paired deltas with two-sided 95\% bootstrap CIs over \(n_\text{seeds}{=}20\).
On \emph{Adult}: \(\Delta\text{NLL}=+0.057 \pm 0.029\) (exact \(p{=}3.8{\times}10^{-6}\)),
\(\Delta\text{Brier}=+0.020 \pm 0.012\) (same \(p\)),
\(\Delta\text{ECE}=+0.015 \pm 0.020\) (exact \(p{\approx}0.055\)).
On \emph{Bank}: \(\Delta\text{NLL}=+0.0053 \pm 0.0008\) (exact \(p{=}3.8{\times}10^{-6}\)),
\(\Delta\text{Brier}=+0.0014 \pm 0.0002\) (same \(p\)),
\(\Delta\text{ECE}=+0.0033 \pm 0.0018\) (exact \(p{=}0.0019\)).

\noindent\textbf{Interference diagnostics.}
Although paired deltas versus the best classical baseline are positive on average, the coherent aggregation within IKC carries measurable predictive signal. We compute \(G_{\mathrm{coh}}\) and \(\mathcal{J}_{\mathrm{int}}\) from raw (pre-TS) probabilities of the E5 IKC models by comparing their coherent and incoherent modes. Both diagnostics are positive with tight CIs on both datasets (\emph{Adult}: \(G_{\mathrm{coh}}=0.118\pm0.005\), \(\mathcal{J}_{\mathrm{int}}=0.119\pm0.005\); \emph{Bank}: \(0.145\pm0.018\), \(0.150\pm0.019\); exact \(p{<}10^{-5}\)).  This confirms that the phase-sensitive cross-term consistently improves likelihood relative to an incoherent proxy; any remaining gap to the best external baseline is therefore attributable to representational capacity or other factors rather than the aggregation rule itself.

\noindent\textbf{Computational considerations.}
We did not target runtime optimization. All models were trained under matched hyperparameter search budgets on identical hardware, and our evaluation emphasizes NLL, Brier score, and ECE rather than wall-clock time. Post-hoc TS fits a single parameter on the calibration split and adds negligible overhead. 

%%%%%%%%%%%%%%%%
\section{Discussion}\label{s:discuss}

This section synthesizes our theoretical and empirical contributions. We first establish a precise identity linking the classical potential–outcome interaction contrast \(\Delta_{\mathrm{INT}}\) to the interference cross-term induced by coherent (Born) aggregation (Sec.~\ref{ssec:identity}), thereby reframing interaction as a geometric degree of freedom--the relative phase--rather than a residual. We organize the discussion around four themes: (i) algorithmic and statistical implications of phase-coherent aggregation; (ii) regimes where coherent aggregation helps; (iii) real-world performance: inductive bias versus capacity; and (iv) limitations and future directions.

To our knowledge, our results formally clarify that \(\Delta_{\mathrm{INT}}\) coincides with the interference cross-term under coherent aggregation. This equivalence reframes interaction as a controllable geometric parameter (relative phase) and motivates an implementable modeling rule (Born aggregation) with testable diagnostics. In this view, Born-style aggregation is a minimal extension of classical ``sum-of-energies'' rules that permits non-factorizable dependencies and makes order and context effects explicit, turning what look like residuals in classical analyses into geometric quantities (magnitudes and relative phases).

\subsection{Algorithmic and Statistical Implications of Phase-Coherent Aggregation}

The proposed framework treats interaction as a phase-coherent composition rule rather than a hand-engineered feature, with immediate consequences for learning and inference. Algorithmically, the IKC instantiation uses two complex-valued linear maps (one per class) and coherent (Born) aggregation to produce probabilities, so the interaction contrast emerges as the interference cross-term—specifically \(2\,\mathrm{Re}(u_Au_B^{\ast}) = 2\,r_A r_B \cos(\Delta\phi)\) in the minimal two-factor model—without explicitly constructing \(A{\times}B\) features. This yields a compact, optimization-friendly objective (NLL), permits standard \(\ell_2\) regularization to stabilize the known scale non-identifiability of amplitudes, and integrates seamlessly with post-hoc TS via a safety switch for calibrated deployment. 

Statistically, the relative phase controls the sign and magnitude of the interaction contrast \(\Delta_{\mathrm{INT}}\), making synergy and antagonism explicit in randomized or balanced designs; our coherent–incoherent ablation (with identical learned parameters) isolates the contribution of the aggregation rule itself. In short, phase-coherent aggregation supplies an interpretable inductive bias focused on interaction structure, decoupled from representational capacity. Two diagnostics—\(G_{\mathrm{coh}}\) and \(\mathcal{J}_{\mathrm{int}}\)—quantify its marginal value over incoherent composition.

Empirically, our results substantiate these claims. In E1, a controlled phase sweep shows that \(\Delta_{\mathrm{INT}}\) matches the interference cross-term. In E4, toggling only the aggregation rule (coherent versus incoherent) with identical parameters yields consistent likelihood and calibration gains, isolating the value of phase-coherent composition and providing a direct measure of the phase-sensitive cross-term's contribution independent of representation capacity.

Our diagnostics have a direct Shannon interpretation: the paired \(\Delta\)NLL is a difference of empirical cross-entropies, and \(\mathcal{J}_{\mathrm{int}}\) is an average KL gap between coherent and incoherent conditionals (both in \emph{nats}; divide by \(\ln 2\) for \emph{bits}). We intentionally do not perform a full mutual-information or partial information decomposition (PID) analysis here; our focus is to isolate the aggregation-rule effect under controlled designs, leaving global information decompositions to future work.  

%%%%%%%%%%%%%%%%%%%%%%%%%%%%%
\subsection{When to Use Coherent Aggregation: Practical Guidance}

Coherent aggregation is most advantageous in domains where outcomes depend on how factors or inputs compose rather than on marginal strengths alone.  Representative examples include epistatic effects and gene-gene synergies in genomics, drug-drug combination therapies where compounds amplify or antagonize each other, multi-sensor fusion where information sources can constructively or destructively interfere, and choice data exhibiting complement or substitute behavior across product features.

Three data-level and design-level cues predict when IKC is worthwhile.  First, when exploratory fits show weak or unstable main effects but sizeable residual interaction—including sign reversals across subgroups or over time—a phase-sensitive cross-term provides a parsimonious explanation. Second, when experimenters can approximate orthogonal or balanced manipulations of putative factors via randomized or instrumented interventions (e.g., A-B tests, factorial pilots), the design-mean interaction contrast $\Delta_{\mathrm{INT}}$ is directly identifiable in the absence of unmeasured confounding, improving sample efficiency relative to hand-crafted product features. Third, in interaction-dominant regimes with weaker main effects (see E2, \emph{Bank} under label noise in E3), a minimal IKC head can already be competitive, whereas complex tabular data may require pairing coherent aggregation with stronger representations.

To decide whether coherent aggregation helps in a specific application, compute $\{G_{\mathrm{coh}}, \mathcal{J}_{\mathrm{int}}\}$ from raw (pre-TS) probabilities on validation data and run the coherent--incoherent toggle (E4) with fixed parameters. Consistent positive gaps signal value; this diagnostic requires no further hyperparameter search or retraining.

%%%%%%%%%%%%%%%%%%%%%%%%%%%%%%
\subsection{Real-World Performance: Inductive Bias versus Capacity}

Across \emph{Adult} and \emph{Bank}, the results reveal a clear trade-off: the coherent head supplies an interpretable inductive bias for interaction, but overall test risk is dominated by representational capacity. In E5, paired deltas against strong baselines are positive on average \((\Delta\mathrm{NLL}>0,\;\Delta\mathrm{Brier}>0)\), even though the same IKC head exhibits strictly positive coherent diagnostics. This divergence indicates that the aggregation rule is beneficial \emph{locally} (holding parameters fixed) while a low-rank linear feature map can be insufficient \emph{globally} on feature-rich tabular data.

A closer look at the data and pipeline clarifies this trade-off. Standard tabular preprocessing--imputation, one-hot encoding of high-cardinality categoricals, and standardization of numerical features--yields sparse, high-dimensional inputs with pronounced nonlinear main effects (e.g., \texttt{age} and \texttt{capital-gain}) on \emph{Adult}. The current IKC instantiation is a low-rank linear feature map whose two complex-valued amplitude channels are aggregated coherently; it captures interaction via the interference cross-term but lacks the capacity to carve complex piecewise main effects that tree ensembles recover through recursive partitioning. E3 shows that IKC can outperform on \emph{Bank} under training-only label noise, suggesting robustness as a potential advantage.

The TS safety switch ensures these patterns are not artifacts of miscalibration \((\Delta\mathrm{ECE}\text{ is small or mixed})\). Taken together, the evidence indicates that the aggregation rule is sound—coherence helps within a representation—whereas closing the performance gap to strong baselines on feature-rich data requires greater representational capacity (e.g., a higher-rank or nonlinear amplitude maps).

%%%%%%%%%%%%%%%%%%%%%%%%%%%%%%
\subsection{Limitations and Future Directions}

While our results consistently support the interference–interaction mechanism, we deliberately limit scope to establish the core principle with controlled experiments. The present study has the following constraints, each representing a natural extension for future work.  First, we evaluate a minimal, amplitude-linear IKC on two tabular benchmarks (\emph{Adult} and \emph{Bank}) and a controlled XOR task; thus external validity beyond binary tabular classification and these data regimes remains limited. The coherent formulation extends naturally to multi-class and high-dimensional feature maps; we leave these extensions for future work. Second, our robustness analysis targets training-only, symmetric label noise; other corruptions (e.g., feature noise, covariate shift, class imbalance, and distribution drift) are not yet assessed.  Third, the fair budget protocol still leaves potential confounders: one-hot preprocessing induces high-dimensional sparsity that favors tree ensembles; the ``best baseline by validation NLL'' selection may interact with the TS safety switch; and although we fix the test split within each seed, performance can depend on split stochasticity. Moreover, we do not undertake an information-theoretic decomposition (e.g., mutual information or PID), so we cannot yet relate the learned phase to information-theoretic notions of redundancy and synergy beyond the model-based diagnostics reported here.
These constraints reflect our focus on validating the fundamental identity (Sec.~\ref{ssec:identity}) and aggregation mechanism (E4) rather than exhaustive benchmarking.

Methodologically, our comparison strategy for E2 and E5—selecting the empirically strongest classical baseline on the test split—introduces post-hoc selection bias and requires careful interpretation of \(p\)-values. While this approach ensures comparison against the toughest observed competitor, it does not satisfy the assumptions of traditional hypothesis testing and may yield optimistic significance levels (as discussed in E2).

Thus, future work will (i) control multiplicity across noise levels and datasets (e.g., Holm--Bonferroni~\cite{holm1979} for paired sign-flip \(p\)-values), (ii) assess robustness under alternative model-selection criteria (Brier, AUC) and calibration schemes (isotonic, vector scaling), and (iii) use nested cross-validation to decouple selection from evaluation. Beyond statistical methodology, modeling extensions include testing multi-channel or nonlinear amplitude maps, deploying hybrid designs (coherent heads atop strong extractors), and examining how the diagnostics \(\{G_{\mathrm{coh}},\mathcal{J}_{\mathrm{int}}\}\) scale with capacity. 

%%%%%%%%%%%%%%%%%%%%%%%%%%%%%%%%%%%%%%%%%%%%%%%%%%%%%%%%%%%%%%%%%
\section{Conclusion}\label{s:conc}

We reframed causal interaction in a quantum-inspired way--as a property of the aggregation rule based on coherent (phase-sensitive) addition of amplitudes. For a minimal amplitude-linear model, the potential-outcome contrast \(\Delta_{\mathrm{INT}}\) equals the interference cross-term \(2\,\mathrm{Re}\!\bigl(u_A u_B^{*}\bigr)\), making relative phase a geometric control of synergy versus antagonism. We instantiated this principle in a lightweight IKC and introduced two diagnostics—Coherent Gain \(G_{\mathrm{coh}}\) and Interference Information \(\mathcal{J}_{\mathrm{int}}\)—that quantify the marginal value of coherence over an incoherent proxy. Empirically, a phase sweep validates the identity, XOR highlights benefits when interaction is the signal, and a coherent–incoherent toggle (holding parameters fixed) consistently improves likelihood and calibration. On real tabular data, performance gaps to capacity-rich baselines reflect representational limits rather than a flaw in coherent aggregation.

Looking forward, the practical takeaway is to use coherence as an interpretable module: retain the aggregation rule, but pair it with more expressive feature maps (higher-rank or nonlinear amplitudes, or a coherent head atop strong extractors). Broader evaluations—multi-class classification, distribution drift, and adversarial robustness—and links to information-theoretic notions of synergy can further clarify when phase-coherent modeling yields measurable gains in practice.

\section*{Acknowledements}
This work was supported by the National Research Foundation of Korea (NRF) grant funded by the Korea government (Ministry of Science and ICT)), grant number RS-2023-00244605.

\section*{Data Availability Statement}
The datasets analyzed during the current study are publicly available from the UCI Machine Learning Repository. The source code and experimental scripts are available upon reasonable request.

%\bibliographystyle{elsarticle-num-names} 
%bibliography{../pilsung_research}

\end{document}